\documentclass{article} 
\usepackage{iclr2020_conference,times}

\usepackage{amsmath,amsfonts,bm}

\def\eqref#1{equation~\ref{#1}}

\def\1{\bm{1}}

\DeclareMathAlphabet{\mathsfit}{\encodingdefault}{\sfdefault}{m}{sl}
\SetMathAlphabet{\mathsfit}{bold}{\encodingdefault}{\sfdefault}{bx}{n}

\newcommand{\softplus}{\zeta}

\DeclareMathOperator*{\argmax}{arg\,max}

\usepackage{latexsym}
\usepackage{url}
\usepackage{graphicx}
\usepackage{algorithm2e}
\usepackage{physics,xfrac,dsfont}
\usepackage{amsmath}
\usepackage{amssymb}
\usepackage{amsfonts}
\usepackage{verbatim}
\usepackage{subcaption}
\usepackage[textsize=scriptsize]{todonotes}

\DeclareMathOperator{\diag}{diag}

\DeclareMathOperator{\enc}{enc}

\DeclareMathOperator{\rnn}{rnn}
\DeclareMathOperator{\att}{att}

\DeclareMathOperator{\layer}{layer}
\DeclareMathOperator{\dense}{dense}

\DeclareMathOperator{\Kuma}{Kuma}

\usepackage{hyperref}
\usepackage{url}

\title{A Latent Morphology Model for Open- \\
Vocabulary Neural Machine Translation}

\author{Duygu Ataman\thanks{Work done while the first author was a doctoral student at the University of Trento and a visiting post-graduate student at the University of Edinburgh.} \\
University of Z\"{u}rich\\
\texttt{ataman@cl.uzh.ch} \\
\And
Wilker Aziz \\
University of Amsterdam \\
\texttt{w.aziz@uva.nl} \\
\And
Alexandra Birch \\
University of Edinburgh \\
\texttt{a.birch@ed.ac.uk}
}

\iclrfinalcopy 
\begin{document}

\maketitle

\begin{abstract}
Translation into morphologically-rich languages challenges neural machine translation (NMT) models with extremely sparse vocabularies where atomic treatment of surface forms is unrealistic. This problem is typically addressed by either pre-processing words into subword units or performing translation directly at the level of characters. The former is based on word segmentation algorithms optimized using corpus-level statistics with no regard to the translation task. The latter learns directly from translation data but requires rather deep architectures. In this paper, we propose to translate words by modeling word formation through a hierarchical latent variable model which mimics the process of morphological inflection. Our model generates words one character at a time by composing two latent representations: a continuous one, aimed at capturing the lexical semantics, and a set of (approximately) discrete features, aimed at capturing the morphosyntactic function, which are shared among different surface forms. Our model achieves better accuracy in translation into three morphologically-rich languages than conventional open-vocabulary NMT methods, while also demonstrating a better generalization capacity under low to mid-resource settings.\\
\end{abstract}

\section{Introduction}

Neural machine translation (NMT) models are conventionally trained by maximizing the likelihood of generating the target side of a bilingual parallel corpus of observations one word at a time conditioned of their full observed context. 
NMT models must therefore learn distributed representations that accurately predict word forms in very diverse contexts, a process that is highly demanding in terms of training data as well as the network capacity.
Under conditions of lexical sparsity, which includes both the case of unknown words and the case of known words occurring in surprising contexts, the model is likely to struggle.
Such adverse conditions are typical of translation involving morphologically-rich languages, where any single root may lead to exponentially many different surface realizations depending on its syntactic context. Such highly productive processes of word formation lead to many word forms being rarely or ever observed with a particular set of morphosyntactic attributes.
The standard approach to overcome this limitation is to 
 pre-process words into 
subword units that are shared among words, which are, in principle, more reliable as they are observed more frequently in varying context \citep{sennrich2016neural,wu2016google}. 
One drawback related to this approach, however, is that the estimation of the subword vocabulary relies on word segmentation methods optimized using corpus-dependent statistics, disregarding any linguistic notion of morphology and the translation objective.
This often produces subword units that are semantically ambiguous as they might be used in far too many lexical and syntactic contexts \citep{ataman2017linguistically}.
Moreover, in this approach, a word form is then generated by prediction of multiple subword units, which makes generalizing to unseen word forms more difficult due to the possibility that a subword unit necessary to reconstruct a given word form may be unlikely in a given context.
To alleviate the sub-optimal effects of using explicit segmentation and generalize better to new morphological forms, recent studies explored the idea of extending NMT to model translation directly at the level of characters \citep{kreutzer2018learning,cherry2018revisiting}, which, in turn, have demonstrated the requirement of using comparably deeper networks, as the network would then need to learn longer distance grammatical dependencies \citep{sennrich2017grammatical}. 


In this paper, we explore the benefits of explicitly modeling variation in surface forms of words using techniques from deep latent variable modeling in order to improve translation accuracy for low-resource and morphologically-rich languages.
Latent variable models allow us to inject inductive biases relevant to the task, which, in our case, is word formation during translation. 
In order to formulate the process of morphological inflection, design a hierarchical latent variable model which translates words one character at a time based on word representations learned compositionally from sub-lexical components. 
In particular, for each word, our model generates two latent representations: i) a continuous-space dense vector aimed at capturing the lexical semantics of the word in a given context, and ii) a set of (approximately) discrete features aimed at capturing that word's morphosyntatic role in the sentence. We then see inflection as decoding a word form, one character at a time, from a learned composition of these two representations.
By forcing the model to encode each word representation in terms of a more compact set of latent features, we encourage them to be shared across contexts and word forms, thus, facilitating generalization under sparse settings.
We evaluate our method in translating English into three morphologically-rich languages each with a distinct morphological typology: Arabic, Czech and Turkish, and show that our model is able to obtain better translation accuracy and generalization capacity than conventional approaches to open-vocabulary NMT. 

\section{Neural Machine Translation}

In this paper, we use recurrent NMT architectures based on the model developed by \citet{bahdanau2014neural}. 
The model essentially estimates the conditional probability of translating a source sequence $x = \langle x_1, x_2, \ldots x_m \rangle$ into a target sequence $y = \langle y_1, y_2, \ldots y_l\rangle$ via an exact factorization:
\begin{equation}\label{nmt1}
\begin{aligned}
p(y|x, \theta) 
    &= \prod_{i=1}^l p(y_j|x, y_{<i}, \theta) 
\end{aligned}
\end{equation}
where $y_{<i}$ stands for the sequence preceding the $i$th target word.
At each step of the sequence, a fixed neural network architecture maps its inputs, the source sentence and the target prefix, to the probability of the $i$th target word observation in context. 
In order to condition on the source sentence fully, this network employs an embedding layer and a bi-directional recurrent neural network (bi-RNN) based encoder. 
Conditioning on the target prefix $y_{<i}$ is implemented using a recurrent neural network (RNN) based decoder, 
and an attention mechanism which summarises the source sentence into a context vector $\mathbf c_i$ as a function of a given prefix \citep{luong2015effective}. 
Given a parallel training set $\mathcal D$, the parameters $\theta$ of the network are estimated to attain a local minimum of the negative log-likelihood function $\mathcal L(\theta|\mathcal D) = -\sum_{x, y \sim \mathcal D}\log p(y|x, \theta)$ via stochastic gradient-based optimization \citep{BottouEtAl2004}.\looseness=-1

\paragraph{Atomic parameterization} estimates the probability of generating each target word $y_{i}$ in a single shot: 
\begin{equation}
\label{eq:nmt2}
p(y_i|x, y_{<i}, \theta) = \frac{\exp(\mathbf E_{y_i}^\top \mathbf h_i)}{\sum_{e=1}^v \exp(\mathbf E_e^\top \mathbf h_i)}  ~,
\end{equation}
where $\mathbf E \in \mathbb R^{v\times d}$ is the target embedding matrix and the decoder output $\mathbf h_i \in \mathbb R^d$ represents $x$ and $y_{<i}$. 
%
%
Clearly, the size $v$ of the target vocabulary plays an important role in determining the complexity of the model, 
which creates an important bottleneck when translating into low-resource and morphologically-rich languages due to the sparsity in the lexical distribution. 

Recent studies approached this problem by performing NMT with \textit{subword} units, a popular one of which is based on the Byte-Pair Encoding algorithm \cite[BPE;][]{sennrich2016neural}, which finds the optimal description of a corpus vocabulary by iteratively merging the most frequent character sequences. 
Atomic parameterization could also be used to model translation at the level of characters, which is found to be advantageous in generalizing to morphological variations \citep{cherry2018revisiting}.

\paragraph{Hierarchical paramaterization} further factorizes the probability of a target word in context:
\begin{equation}
   p(y_i|x, y_{<i}, \theta) 
   = \prod_{j=1}^{l_i}  p(y_{i,j}|x, y_{<i}, y_{i,<j}, \theta) 
\end{equation}
where the $i$th word $y_i = \langle y_{i,1}, \ldots, y_{i, l_i}\rangle$ is seen as a sequence of $l_i$ characters. Generation follows one character at a time, each with probability computed by a fixed neural network architecture with varying inputs, namely, the source sentence $x$, the target prefix $y_{<i}$, and the prefix $y_{i,<j}$ of characters already generated for that word.
In this case there are two recurrent cells, one updated at the boundary of each token, much like in the standard case, and another updated at the character level. 
\cite{luong-hybrid} propose hierarchical parameterization to compute the probability $p(y_i|x, y_{<i}, \theta)$ for unknown words, while for known words they use the atomic parameterization.
In this paper, we use the hierarchical parameterization method for generating \emph{all} target words, where we also augment the input embedding layer with a character-level bi-RNN, which computes each word representation $\mathbf y_i$ as a composition of the embeddings of their characters \citep{DBLP:journals/corr/LingTDB15}.

\section{A Latent Morphology Model (LMM) for Learning Word Representations}
The application of a hierarchical structure for learning word representations in language modeling \citep{vania2017characters} or semantic role labeling \citep{sahin2018character} have shown that such representations encode many cues about the morphological features of words by establishing a mapping between phonetic units and lexical context. Although it can provide an alternative solution to open-vocabulary NMT by potentially alleviating the need for subword segmentation, the quality of word representations learned by an hierarchical model is still highly dependant on the amount of observations \citep{sahin2018character,ataman2019importance}, since the training data is essential in properly modeling the lexical distribution.
On the other hand, the process of word formation, particularly morphological inflection, has many properties that remain universal across languages, where a word is typically composed of a \emph{lemma}, representing its lexical semantics, and a distinct combination of categorical \textit{inflectional features} expressing the word's syntactic role in the phrase or sentence. In this paper, we propose to manipulate this universal structure in order to enforce an inductive bias on the prior distribution of words and allow the hierarchical parameterization model in properly learning lexical representations under conditions of data sparsity. 

\subsection{\label{sec:genmodel}Generative model}

Our generative LMM for NMT formulates word formation in terms of a stochastic process, where each word is generated one character at a time by composing two latent representations: 
a continuous vector aimed at capturing the lemma, and a set of sparse features aimed at capturing the inflectional features.
The motivation for using a stochastic model is twofold. First, deterministic models are by definition \emph{unimodal}: when presented with the same input (the same context) they always produce the same output. When we model the word formation process, it is reasonable to expect a larger degree of ambiguity, that is, for the same context (\textit{e.g.} a noun prefix), we may continue by inflecting the word differently depending on the (latent) mode of operation we are at (\textit{e.g.} generating nominative, accusative or dative noun). 
Second, in stochastic models, the choice of distribution gives us a mechanism to favour a particular type of representation. In our case, we use \emph{sparse} distributions for inflectional features to accommodate the fact that morphosyntactic features are discrete in nature. 
Our latent variable model is an instance of a variational auto-encoder \citep[VAE;][]{Kingma+2014:VAE} inspired by the model of \cite{zhou2017multi} for morphological reinflection.

\begin{figure}[t]
    \centering
    \includegraphics[scale=0.085]{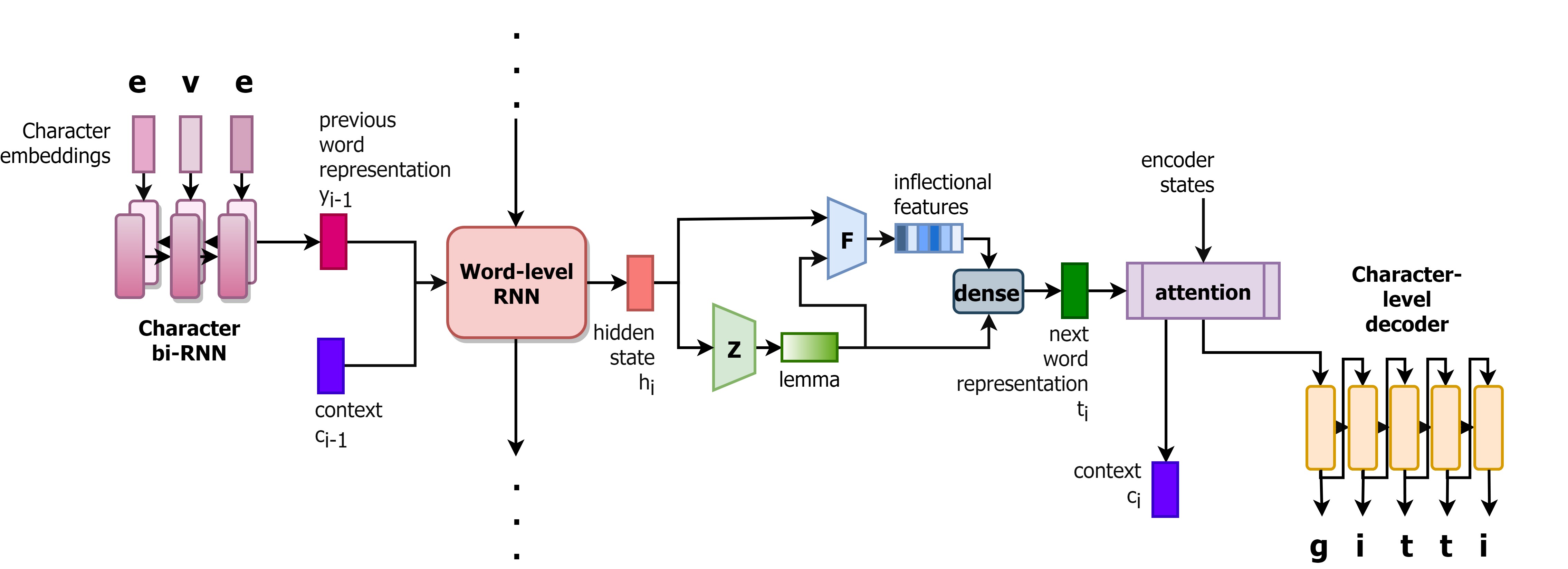}
    \caption{LMM for computing word representations while translating the sentence \textit{`... went home'} into Turkish (\textit{`eve}-\textbf{(to)home} \textit{gitti}\textbf{(he/she/it)went'}). The character-level decoder is initialized with the attentional vector $\mathbf h_i$ computed by the attention mechanism using current context $\mathbf c_i$ and the word representation $\mathbf t_i$ as in \cite{luong-hybrid}.}
    \label{fig:cvae}
\end{figure}

Generation of the $i$th word starts by sampling a Gaussian-distributed representation in context. This requires predicting the Gaussian location $\mathbf u_i$ and scale $\mathbf s_i$ vectors,\footnote{\textbf{Notation} ~ We use capital Roman letters for random variables (and lowercase letters for assignments). Boldface Roman letters are reserved for neural network output vectors, and $\odot$ stands for elementwise multiplication. Finally, we denote typical neural network layers as $\layer(\text{inputs};\text{parameters})$.}
\begin{equation}
\begin{aligned}
    Z_i|x, y_{<i} &\sim \mathcal N(\mathbf u_i, \diag(\mathbf s_i \odot \mathbf s_i)) \\ 
    \mathbf u_i &= \dense(\mathbf h_i; \theta_{\text{u}}) \\
    \mathbf s_i &= \softplus(\dense(\mathbf h_i; \theta_{\text{s}})) 
\end{aligned}
\end{equation}
where prediction of the location (in $\mathbb R^d$) and scale (in $\mathbb R^d_{>0}$) from the word-level decoder hidden state $\mathbf h_i$ (which represents $x$ and $y_{<i}$) is performed by two dense layers, and the scale values are ensured to be positive with the softplus ($\softplus$) activation.\footnote{In practice, we sample $z_i$ via a reparameterization in terms of a fixed Gaussian, namely, $z_i = \mathbf u_i + \epsilon_i \odot \mathbf s_i$ for $\epsilon_i \sim \mathcal N(0, I_d)$. This is known as the \emph{reparameterization trick} \citep{Kingma+2014:VAE}, which allows back-propagation through stochastic units \citep{RezendeEtAl14VAE}.}

Generation proceeds by then sampling a $K$-dimensional vector $f_i$ of sparse scalar features (see \S\ref{sec:sparse}) conditioned on the source $x$, the target prefix $y_{<i}$, and the sampled lemma $z_{i}$. We model sampling of $f_i$ conditioned on $z_i$ in order to capture the insight that inflectional transformations typically depend on the category of a lemma.
Having sampled $f_i$ and $z_i$, the representation of the $i$th target word is computed by a transformation of $z_i$ and $f_i$, 
i.e. $\mathbf t_i = \dense([z_i, f_i]; \theta_{\text{comp}})$.

As shown in Figure \ref{fig:cvae}, our model generates each word character by character  auto-regressively by conditioning on the word representation $\mathbf t_i$ predicted by the LMM, the current context $\mathbf c_i$, and the previously generated characters 
following the hierarchical parameterization.\footnote{Formally, because the decoder is an RNN, we are also conditioning on $z_{<i}$ and $f_{<i}$. We omit this dependence to avoid clutter.}
See Algorithm \ref{alg:generation} for details on generation.

\begin{algorithm}[H]
 \KwIn{model parameters $\theta$, latent lemma $z_i$, latent morphological attributes $f_i$, observed character sequence $\langle y_{i,1}, \ldots, y_{i,l_i} \rangle$ if training or placeholders if test, decoder state $\mathbf h_i$, and context vector $\mathbf c_i$}
 \KwResult{updated decoder hidden state, prediction (a word), probability of prediction (for loss)}
 initialization\;
 $\mathbf t_i = \dense([z_i, f_i];\theta_{\text{comp}})$ \;
 initialize char-rnn with a projection of $[\mathbf t_i, \mathbf c_i]$\;
 \For{$j < l_i$ and $j < \text{max}$}{ 
    compute output layer from char-rnn state \;
    \eIf{training}{
        set prediction to observation ${y}_{i,j}$ \;
    }{
        set prediction to $\argmax$ of output softmax layer \;
    }
    assess log-probability of prediction \;
    update word-level RNN decoder with prediction\;
 }
 \caption{\label{alg:generation}Word generation: in training the word is observed, thus we only update the decoder and assess the probability of the observation, in test, we use mean values of the distributions to represent most likely values for $z$ and $f$ and populate predictions with beam-search.}
\end{algorithm}







\subsection{\label{sec:sparse}Sparse features}

Since each target word $y_i$ may have multiple inflectional features, ideally, we would like $f_i$ to be $K$ feature indicators, which could be achieved by sampling from $K$ independent Bernoulli distributions parameterized in context. 
The problem with this approach is that sampling Bernoulli outcomes is non-differentiable, thus, their training requires gradient estimation via REINFORCE \citep{Williams1992} and sophisticated variance reduction techniques. 
An alternative approach that has recently become popular is to use relaxations such as the Concrete distribution or Gumbel-Softmax \citep{MaddisonEtAl2017:Concrete,JangEtAl2017:GumbelSoftmax} in combination with the straight-through estimator \citep[ST;][]{bengio2013estimating}.  
This is based on the idea of relaxing the discrete variable from taking on samples in the discrete set $\{0,1\}$ to taking on samples in the continuous set $(0, 1)$ using a distribution for which a reparameterization exists (\textit{\textit{e.g.}} Gumbel). Then, a non-differentiable activation (\textit{\textit{e.g.}} a threshold function) maps continuous outcomes to discrete ones. 
ST simply ignores the discontinuous activation in the backward pass, \textit{\textit{i.e.}} it assumes the Jacobian is the identity matrix. 
This does lead to biased estimates of the gradient of the loss, which is in conflict with the requirements behind  stochastic optimization \citep{robbinsmonro:1951}.

An alternative presented by \citet{louizos2017learning} achieves a different compromise, it gets rid of bias at the cost of mixing both sparse and dense outcomes. 
The idea is to obtain a continuous sample $c \in (0, 1)$ from a distribution for which a reparameterization exists and stretch it to a continuous support $(l, r) \supset (0, 1)$ using a simple linear transformation $s=l + (r-l) c$. 
A rectifier is then employed to map the negative outcomes to $0$ and the positive outcomes larger than one to $1$, \textit{i.e.} $f = \min(1, \max(0, s))$. The rectifier is only non-differentiable at $s=0$ and at $s=1$, however, because the stretched variable $s$ is sampled from a \emph{continuous} distribution, the chance of sampling $s=0$ and $s=1$ is essentially $0$. 
This stretched-and-rectified distribution allows: i) the sampling procedure to become differentiable with respect to the parameters of the distribution, ii) to sample sparse outcomes with an unbiased estimator, 
and iii) to calculate the probability of sampling $f=0$ and $f=1$ in closed form as a function of the parameters of the underlying distribution, which corresponds to the probability of sampling $s<0$ and $s>1$, respectively. 

In their paper, \citet{louizos2017learning} used the BinaryConcrete (or Gumbel-Sigmoid) as the underlying continuous distribution, the sparsity of which is controlled via a temperature parameter. However, in our study, we found this parameter difficult to predict, since it is very hard to allow a neural network to control its value without unstable gradient updates. Instead, we opt for a slight variant by \citet{bastings2019} 
based on the Kumaraswamy distribution \citep{kumaraswamy1980generalized}, a two-parameters distribution that closely resembles a Beta distribution and is sparse whenever its (strictly positive) parameters are between $0$ and $1$. In the context of text classification, \citet{bastings2019} shows this stretch-and-rectify technique to work better than methods based on REINFORCE.

For each token $y_i$, we sample $K$ independent Kumaraswamy variables in context,
\begin{equation}
\begin{aligned}
    &C_{i,k}|x, y_{<i}, z_i \sim \Kuma(a_{i,k}, b_{i,k}) \quad k=1, \ldots, K\\
    &\quad [\mathbf a_i, \mathbf b_i] = \softplus(\dense([z_i, \mathbf h_i]; \theta_{\text{ab}})) 
\end{aligned}
\end{equation}
which makes a continuous random vector $c_i$ in the support $(0, 1)^K$.\footnote{In practice we sample $c_{i,k}$ via a reparameterization of a fixed uniform variable, namely, $c_{i,k} =  (1- (1-\varepsilon_{i,k})^{\sfrac{1}{b_{i,k}}})^{\sfrac{1}{a_{i,k}}}$ where $\varepsilon_{i,k} \sim \mathcal U(0, 1)$, which much like the Gaussian reparameterization enables back-propagation through samples \citep{nalisnick2017stick}.}
We then stretch-and-rectify the samples via 
$f_{i,k} = \min(1, \max(0, l - (r - l) c_{i,k}))$
making $f_i$ a random vector in the support $[0, 1]^K$.\footnote{We use $l=-0.1$ and $r=1.1$. Figure \ref{fig:kuma} in the appendix illustrates different instances of this distribution.}
The probability that $f_{i,k}$ is exactly $0$ is
\begin{subequations}\label{eq:pi}
\begin{equation}
    \pi_{i,k}^{\{0\}} = \int_0^{\frac{-l}{r-l}} \Kuma(c|a_{i,k}, b_{i,k}) \dd c
\end{equation}
and the probability that $f_{i,k}$ is exactly $1$ is 
\begin{equation}
    \pi_{i,k}^{\{1\}} = 1 - \int_0^{\frac{1-l}{r-l}} \Kuma(c|a_{i,k}, b_{i,k}) \dd c
\end{equation}
and therefore the complement
\begin{equation}
\pi_{i,k}^{(0,1)} = 1 - \pi_{i,k}^{\{0\}} - \pi_{i,k}^{\{1\}} 
\end{equation}
\end{subequations}
is the probability that $f_{i,k}$ be any continuous value in the open set $(0, 1)$.
In \S\ref{sec:reg}, we will derive regularizers based on $\pi_{i,k}^{(0,1)}$ to promote sparse outcomes to be sampled with large probability.


\subsection{Parameter estimation}

Parameter estimation of neural network models is typically done via maximum-likelihood estimation (MLE), where we approach a local minimum of the negative log-likelihood function via stochastic gradient descent with gradient computation automated by the back-propagation algorithm. 
Using the following shorthand notation:
\begin{subequations}
\begin{align}
    \alpha(z_i) &\triangleq p(z_i|x, y_{<i}, z_{<i}, f_{< i}, \theta) \\
    \beta(f_i) &\triangleq \prod_{k=1}^K p(f_{i,k}|x, y_{<i}, z_{<i}, f_{<i}, z_i, \theta) \\
    \gamma(y_i) &\triangleq  \prod_{j=1}^{l_i} p(y_{i,j}|x, y_{<i}, z_{\le i}, f_{\le i},  y_{i, <j}, \theta) ~.
\end{align}
\end{subequations} 
The log-likelihood for a single data point can be formulated as:
\begin{equation}
\log p(y|x, \theta) = \log \int \prod_{i=1}^l \alpha(z_i)\beta(f_i)\gamma(y_i) \dd z \dd f
\end{equation}
the computation of which is intractable. Instead, we resort to variational inference \citep[VI;][]{Jordan+1999:VI}, where we optimize a lower-bound on the log-likelihood 
\begin{equation}
    \mathbb E_{q(z, f|x,y, \lambda)}\left[ \sum_{i=1}^l \log \frac{  \alpha(z_i)\beta(f_i)\gamma(y_i)}{q(z,f|x, \lambda)}\right] 
\end{equation}
expressed with respect to an independently parameterized posterior approximation $q(z, f|x, y, \lambda)$. For as long as sampling from the posterior is tractable and can be performed via a reparameterization, we can rely on stochastic gradient-based optimization. 
In order to have a compact parameterization, we choose 
\begin{equation}
    q(z, f|x, y, \lambda) := \prod_{i=1}^l \alpha(z_i) \beta(f_i) ~.
\end{equation}
This simplifies the lowerbound, which then takes the form of $l$ nested expectations, the $i$th of which is
$\mathbb E_{\alpha(z_i)\beta(f_i)}\left[ \log \gamma(y_i)  \right]$.
This is similar to the stochastic decoder of \citet{schulz-etal-2018-stochastic}, though our approximate posterior is in fact, also our parameterized prior.\footnote{This means we refrain from conditioning on the observation $y_i$ itself when sampling $z_i$ and $f_i$. Whereas this gives our posterior approximation access to less features than the true posterior would have, we do not employ a fixed uninformative prior, but rather an autoregressive network trained on the likelihood of generated word forms. This is a common approximation for latent variable models that employ autoregressive priors \citep{goyal2017z}.} 
Although this objective does not particularly promote sparsity, we employ sparsity-inducing regularization techniques that will be discussed in the next section.

Concretely, for a given source sentence $x$, target prefix $y_{<i}$, and a latent sample $z_{\le i}, f_{\le i}$, we obtain a single-sample estimate of the loss by computing $\mathcal L_i(\theta) = - \log \gamma(y_i)$.

\subsection{Regularization}
\label{sec:reg}

In order to promote sparse distributions for the inflectional features, we apply a regularizer inspired by expected $L_0$ regularization \citep{louizos2017learning}. Whereas $L_0$ is a penalty based on the number of non-zero outcomes, we design a penalty based on the expected number of \emph{continuous outcomes}, which corresponds to  $\pi_{i,k}^{(0,1)}$ as shown in Equation (\ref{eq:pi}).
For a given source sentence $x$, target prefix $y_{<i}$, and a latent sample $z_{< i}, f_{ < i}$, we aggregate this penalty for each feature
\begin{equation}
    \mathcal R_i(\theta) = \sum_{k=1}^{K} \pi_{i,k}^{(0,1)}
\end{equation}
and add it to the cost function with a positive weight $\rho$. 
The final loss of the NMT model is 
\begin{equation}
   \mathcal L(\theta|\mathcal D) = \sum_{x,y\sim \mathcal D}\sum_{i=1}^{|y|} \mathcal L_i(\theta) + \rho \mathcal R_i(\theta) ~.
\end{equation}

\subsection{Predictions}

In our model, obtaining the conditional likelihood for predicting the most likely hypothesis requires marginalisation of the latent variables, which is intractable.
An alternative approach is to heuristically search through the joint distribution, 
\begin{equation}
    \argmax_{y, z, f} ~ p(y, z, f|x) ~,
\end{equation}
rather than the marginal, an approximation that has been referred to as \emph{Viterbi decoding} \citep{smith2011linguistic}.
During beam search, we populate the beam with alternative target words, and for each prefix $y_{<i}$ in the beam, we resort to deterministically choosing the latent variables based on a single sample which we deem representative of their distributions, which is a common heuristic in VAEs for translation \citep{zhang-etal-2016-variational-neural,schulz-etal-2018-stochastic}.
For unimodal distributions, such as the Gaussian $p(z_i|x, y_{<i}, z_{<i}, f_{<i})$, we use the analytic mean, whereas for multimodal distributions, such as the Hard Kumaraswamy $p(f_i|x, y_{<i}, z_{\le i}, f_{<i})$, we use the argmax.\footnote{We maximize across the three configurations of each feature, namely, $\max \{ \pi_{i,k}^{\{0\}}, \pi_{i,k}^{\{1\}}, \pi_{i,k}^{(0, 1)} \}$. If $\pi_{i,k}^{(0, 1)}$ is highest, we return the mean of the underlying Kumaraswamy variable.}

\section{Evaluation}

\subsection{Models}

We evaluate our model by comparing it in machine translation against three baselines which constitute the conventional open-vocabulary NMT methods, including architectures using atomic parameterization either with subword units segmented with BPE \citep{sennrich2016neural} or characters, and the hierarchical parameterization method employed for generating all words in the output. We implement all architectures using Pytorch \citep{paszke2017automatic} within the OpenNMT-py framework \citep{klein2017opennmt}\footnote{Our software is available at: \texttt{https://github.com/d-ataman/lmm}}. 

\subsection{Data and Languages}

In order to evaluate our model we design two sets of experiments. The experiments in \S\ref{subsec:morphtyp} aim to evaluate different methods under low-resource settings, for languages with different morphological typology. We model the machine translation task from English into three languages with distinct morphological characteristics: Arabic (\textit{templatic}), Czech (\textit{fusional}), and Turkish (\textit{agglutinative}). We use the TED Talks corpora \citep{mauro2012wit3} for training the NMT models for these experiments. In \S\ref{subsec:datasize}, we conduct more experiments in Turkish to demonstrate the case of increased data sparsity using multi-domain training corpora, where we extend the training set using corpora from EU Bookshop \citep{skadicnvs2014billions}, Global Voices, Gnome, Tatoeba, Ubuntu \citep{tiedemann2012parallel}, KDE4 \citep{tiedemann2009news}, Open Subtitles \citep{lison2016opensubtitles2016} and SETIMES \citep{tyers2010south}\footnote{The size of the resulting combined corpora is further reduced to filter out noise and reduce the computational cost of the experiments using data selection methods \citep{cuong2014latent}.}. The statistical characteristics of the training sets are given in Tables \ref{tab:data1} and \ref{tab:data2}. We use the official evaluation sets of the IWSLT\footnote{The International Workshop on Spoken Language Translation} for validating and testing the accuracy of the models. In order to increase the number of unknown and rare words in the evaluation sets we measure accuracy on large test sets combining evaluation sets from many years (Table \ref{tab:devtest} presents the evaluation sets used for development and testing). The accuracy of each model output is measured using BLEU \citep{bleu} and chrF3 \citep{popovic2015chrf} metrics, whereas the significance of the improvements are computed using bootstrap hypothesis testing \citep{clark2011better}. In order to measure the accuracy in predicting the correct syntactic description of the references, we also compute BLEU scores over the output sentences segmented using a morphological analyzer. We use the AlKhalil Morphosys \citep{boudchiche2017alkhalil} for segmenting Arabic, Morphidata \citep{strakova14} for segmenting Czech and the morphological lexicon model of Oflazer \citep{oflazer1994tagging} and disambiguation tool of Sak \citep{sak2007morphological} for segmenting Turkish sentences into sequences of lemmas and morphological features.

\subsection{Training Settings}

All models are implemented using gated recurrent units (GRU) \citep{cho2014properties}, and have a single-layer bi-RNN encoder. 
The source sides of the data used for training all NMT models, and the target sides of the data used in training the subword-level NMT models are segmented using BPE with 16,000 merge rules. 
We implement all decoders using a comparable number of GRU parameters, including 3-layer stacked-GRU subword and character-level decoders, where the attention is computed after the $1^{st}$ layer \citep{barone2017deep} and a 3-layer hierarchical decoder which implements the attention mechanism after the $2^{nd}$ layer. 
All models use an embedding dimension and GRU size of 512. 
LMM uses the same hierarchical GRU architecture, where the middle layer is augmented using 4 multi-layer perceptrons with 256 hidden units. We use a lemma vector dimension of 150, 10 inflectional features (See \S\ref{subsec:featuredim} for experiments conducted to tune the feature dimensions) and set the regularization constant to $\rho=0.4$. All models are trained using the Adam optimizer \citep{kinga2015method} with a batch size of 100, dropout rate of 0.2, learning rate of 0.0004 and learning rate decay of 0.8, applied when the perplexity does not decrease at a given epoch.\footnote{Perplexity is the exponentiated average negative log-likelihood per segment (BPE, or character) that a model assigns to a dataset. It corresponds to the model's average surprisal per time step.} Translations are generated with beam search with a beam size of 5, where the hierarchical models implement the hierarchical beam search algorithm \citep{ataman2019importance}.

\subsection{Results}

\subsubsection{The Effect of Morphological Typology}
\label{subsec:morphtyp}

The experiment results given in Table \ref{tab:results1} show the performance of each model in translating English into Arabic, Czech and Turkish. In Turkish, the most sparse target language in our benchmark with rich agglutinative morphology, using character-based decoding shows to be more advantageous compared to the subword-level and hierarchical models, suggesting that increased granularity in the vocabulary units might aid in better learning accurate representations under conditions of high data sparsity.
In Arabic, on the other hand, using a hierarchical decoding model shows to be advantageous compared to the subword and character-level models, as it might be useful in better learning syntactic dependencies.
LMM obtains improvements of {\bf 0.51} and {\bf 0.30} BLEU points in Arabic and Turkish over the best performing baselines, respectively. 
The fact that our model can efficiently work in both Arabic and Turkish confirms that it can handle the generation of both concatenative and non-concatenative morphological transformations.
The results in the English-to-Czech translation direction do not indicate a specific advantage of using either method for generating fusional morphology, where morphemes are already optimized at the surface level, although our model is still able to achieve  translation accuracy comparable to the character and subword-level models.

\begin{table*}[t]
\centering
\begin{small}
\begin{tabular}{c|ccc|ccc|ccc}
 & \multicolumn{9}{c}{\bf \it (only in-domain)} \\
    {\bf Model} & \multicolumn{3}{c|}{\bf AR} & \multicolumn{3}{c|}{\bf CS} & \multicolumn{3}{c}{\bf TR}  \\
    & {\bf BLEU}& {\bf t-BLEU} & {\bf chrF3} & {\bf BLEU}& {\bf t-BLEU} & {\bf chrF3} & {\bf BLEU}& {\bf t-BLEU} & {\bf chrF3}\\
      \hline
Subwords & 14.27 & 51.24 & 0.3927 &  16.60 & \bf 54.22 & \bf 0.4123 &  8.52 & 38.03 & 0.3763 \\ 
\hline
Char.s & 12.72 & 47.56 & 0.3804 & 16.94  & 52.80 & 0.4103 & 10.63 & 40.63 & 0.3810 \\
\hline
Hierarch. & 15.55 & 54.01 & 0.4154 & 16.79 & 48.27& 0.4068 & 9.74 & 35.91 & 0.3771 \\
\hline
LMM & \bf 16.06 & \bf 55.97 & \bf 0.4251 & \bf 16.97 & 50.35  &  0.4095 &  \bf 10.93 & \bf 45.47 & \bf 0.3889 \\
\vspace{0.1cm}
\end{tabular}
\quad
\begin{tabular}{c|ccc}
 & \multicolumn{3}{c}{\bf \it (multi-domain)}\\
   {\bf Model} & \multicolumn{3}{c}{\bf TR}  \\
   & {\bf BLEU} & {\bf t-BLEU} & {\bf chrF3} \\
      \hline
Subwords &  10.42 & 42.65 & 0.3722 \\ 
\hline
Char.s &  8.94 & 37.12 & 0.3274 \\ 
\hline
Hierarch. & 10.35 & 40.54 & 0.3870 \\ 
\hline
LMM  & \bf 11.48 & \bf 48.23 & \bf 0.3939 \\ 
\end{tabular}
\caption{Above: Machine translation accuracy in Arabic (AR), Czech (CS) and Turkish (TR) in terms of BLEU and ChrF3 metrics as well as BLEU scores computed on the output sentences tagged with the morphological analyzer (t-BLEU) using in-domain training data. Below: The performance of models trained with multi-domain data. Best scores are in bold. All improvements over the baselines are statistically significant (p-value~$<$~0.05).}
\label{tab:results1}
\end{small}
\end{table*}

\subsubsection{Predicting Unseen Words}

In addition to the general machine translation evaluation using automatic metrics, we perform a more focused statistical analysis to illustrate the performance of different methods in predicting unseen words by computing the average perplexity per character on the input sentences which contain out-of-vocabulary (OOV) words as suggested by \citet{cotterell-etal-2018-languages}. We also analyze the outputs generated by each decoder in terms of the frequency of unknown words in each model output and the Kullback-Leibler (KL) divergence between the character trigram distributions of the references and outputs, which represents the coherence between the statistical distribution learned by each model and the reference translations. 

Our analysis results generally confirm the advantage of increased granularity during the generation of unseen words, where the character-level decoder can generate a higher rate of unseen word forms and higher KL-divergence with the reference, suggesting superior ability in generalizing to new output and not necessarily copying previous observations as the subword-level model, however, this advantage is more visible in Turkish and less in Czech or Arabic. The hierarchical decoder which performs the search at the level of words, on the other hand, behaves with less uncertainty in terms of the perplexity values although it cannot demonstrate the ability to generalize to new forms and neither can closely capture the actual distribution in the target language.

Due to its stochastic nature, our model yields higher perplexity values compared to the hierarchical model, whereas the values range between subword and character-based models, possibly finding an optimal level of granularity between the two solutions. The KL-divergence and OOV rates confirm that our model has the potential in better generalize to new word forms as well as different morphological typology.

\begin{table}[h]
\centering
\begin{small}
\begin{tabular}{c|ccc|ccc|ccc}
    {\bf Model} & 
    \multicolumn{3}{c}{\bf AR} & \multicolumn{3}{c}{\bf CS} & \multicolumn{3}{c}{\bf TR} \\
    & \bf OOV\% & \bf Ppl & \bf KL-Div  & \bf OOV\% & \bf Ppl & \bf KL-Div  & \bf OOV\% & \bf Ppl & \bf KL-Div  \\
      \hline
Subwords & 1.75 & 2.84 & 12,871 &2.39& 2.62 & 8,954 & 3.54 & 2.78 & 17,342 \\
\hline
Char.s &3.08& 2.46 & 29,607 & 1.90& 2.61 & 17,092 & 4.28& 2.38 & 38,043\\
\hline
Hierarch. & 1.96 & 2.59 & 15,064 &0.87& 2.65 & 29,022 & 1.53 & 2.46 & 68,743\\
\hline
LMM &3.78 & 2.68 & 9,892 &2.4& 2.71 & 14,296 & 4.89& 2.59 & 38,930\\
\end{tabular}
\end{small}
\caption{Percentage of out-of-vocabulary (OOV) words in the output, normalized perplexity measures (PPl) per characters and the KL divergence between the reference and outputs of systems trained with in-domain data on different language directions.}
\label{tab:results3}
\end{table}

\subsubsection{The Effect of Data Size}
\label{subsec:datasize}

Repeating the experiments in the English-to-Turkish translation direction by increasing the amount of training data with multi-domain corpora demonstrates a more challenging case, where there is a greater possibility of observing new words in varying context, either in the form of morphological inflections due to previously unobserved syntactic conditions, or a larger vocabulary extended with terminology from different domains. In this experiment, the character-level model experiences a drop in performance and its accuracy is much lower than the subword-level one, suggesting that its capacity cannot cope with the increased amount of training data.
Empirical results suggest that with increased capacity, character-level models carry the potential to reach comparable performance to subword-level models \citep{cherry2018revisiting}. On the other hand, our model reaches a much larger improvement of {\bf 0.82} BLEU points over the subword-level and {\bf 2.54} BLEU points over the character-level decoders, suggesting that it could make use of the increased amount of observations for improving the translation performance, which possibly aid the morphology model in becoming more accurate. 

\subsubsection{The Impact of Inflectional Features}

In order to understand whether the latent inflectional features in fact capture information about variations related to morphological transformations, we first try generating different surface forms of the same lemma by sampling a lemma vector with LMM for the input word \textit{'go'} and generating outputs using the fixed lemma vector and assigning different values to the inflectional features. In the second experiment, we assess the impact of the inflectional features by setting all features $f$ to 0 and translating a set of English sentences with varying inflected forms in Turkish. Table \ref{tab:chap8.differentoutputs} presents different sets of feature values and the corresponding outputs generated by the decoder and the outputs generated with or without the inflectional component.

\begin{table}[!h]
    \centering
    \begin{small}
    \begin{tabular}{c|c|c}
        \bf Features & \bf Output & \bf English Translation\\
        \hline
        $[$1,1,1,1,1,1,1,1,1,1$]$ & git &  go \it (informal)\\
        $[$0,1,1,1,1,1,1,1,1,1$]$ & '{\it a} git & {\it to} go\\
        $[$0,1,0,1,1,1,1,1,1,1$]$ & '{\it da} git & {\it at} go\\
        $[$0,0,0,1,1,0,0,1,1,0$]$ & gidin & go \it (formal)\\
        $[$1,1,0,0,0,0,1,0,1,1$]$ & gitmek & to go \it (infinitive) \\
        $[$0,0,1,0,0,0,0,0,0,1$]$ & gidiyor  & (he/she/it is) going\\
        $[$0,0,0,0,0,0,0,0,1,0$]$ & gidip &  {\it by} going \it (gerund)\\
        $[$0,0,1,1,0,0,1,0,1,0$]$ & gidiyoruz & (we are) going\\
        \vspace{0.1cm}
    \end{tabular}
    \quad
\begin{tabular}{c|c|c}
\bf Input & \bf Output with $f$ & \bf Output without $f$\\
\hline
    he went home. & ev\textbf{e} gitti.& ev\textbf{e} gitti.\\
    he came from home. & ev\textbf{den} geldi. &ev\textbf{e} geldi.\\
    it is good to be home. & ev\textbf{de} olmak iyi. & ev\textbf{de} olmak iyi. \\
    his home has red walls. & ev\textbf{inde} k{\i}rm{\i}z{\i} duvar\textbf{lar} var. & ev\textbf{de} k{\i}rm{\i}z{\i} duvar var.\\
    
\end{tabular}
    \end{small}
    \caption{Above: Outputs of LMM based on the lemma \textit{`git'} (\textit{`go'}) and different sets of inflectional features. Below: Examples of predicting inflections in context with or without using features.}
    \label{tab:chap8.differentoutputs}
\end{table}

The model generates different surface forms for different sets of features, confirming that the latent variables represent morphological features related to the infinitive form of the verb, as well as its formality conditions, prepositions, person, number and tense. 
Decoding the set of sentences given in the second experiment LMM always generates the correct inflectional form, although when the feature values are set to 0 the model omits some inflectional features in the output, suggesting that despite partially relying on the source-side context, it still encodes important information for generating correct surface forms in the inflectional features.

\section{Conclusion}

In this paper we presented a novel decoding architecture for NMT employing a hierarchical latent variable model to promote sparsity in lexical representations, which demonstrated promising application for morphologically-rich and low-resource languages. Our model generates words one character at a time by composing two latent features representing their lemmas and inflectional features. We evaluate our model against conventional open-vocabulary NMT solutions such as subword and character-level decoding methods in translationg English into three morphologically-rich languages with different morphological typologies under low to mid-resource settings. Our results show that our model can significantly outperform subword-level NMT models, whereas demonstrates better capacity than character-level models in coping with increased amounts of data sparsity. We also conduct ablation studies on the impact of feature variations to the predictions, which prove that despite being completely unsupervised, our model can in fact manage to learn morphosyntactic information and make use of it to generalize to different surface forms of words.

\section{Acknowledgments}

The authors would like to thank Marcello Federico, Orhan F{\i}rat, Adam Lopez, Graham Neubig, Akash Srivastava and Clara Vania for their feedback and suggestions. This project received funding from the European Union’s Horizon 2020 research and innovation programme under grant agreements 825299 (GoURMET) and 688139 (SUMMA).



\bibliography{iclr2020_conference}

\begin{thebibliography}{51}
\providecommand{\natexlab}[1]{#1}
\providecommand{\url}[1]{\texttt{#1}}
\expandafter\ifx\csname urlstyle\endcsname\relax
  \providecommand{\doi}[1]{doi: #1}\else
  \providecommand{\doi}{doi: \begingroup \urlstyle{rm}\Url}\fi

\bibitem[Ataman et~al.(2017)Ataman, Negri, Turchi, and
  Federico]{ataman2017linguistically}
Duygu Ataman, Matteo Negri, Marco Turchi, and Marcello Federico.
\newblock Linguistically-motivated vocabulary reduction for neural machine
  translation from {T}urkish to {E}nglish.
\newblock \emph{The Prague Bulletin of Mathematical Linguistics}, 108\penalty0
  (1):\penalty0 331--342, 2017.

\bibitem[Ataman et~al.(2019)Ataman, Firat, Di~Gangi, Federico, and
  Birch]{ataman2019importance}
Duygu Ataman, Orhan Firat, Mattia~A Di~Gangi, Marcello Federico, and Alexandra
  Birch.
\newblock On the importance of word boundaries in character-level neural
  machine translation.
\newblock \emph{arXiv preprint arXiv:1910.06753}, 2019.

\bibitem[Bahdanau et~al.(2014)Bahdanau, Cho, and Bengio]{bahdanau2014neural}
Dzmitry Bahdanau, Kyunghyun Cho, and Yoshua Bengio.
\newblock Neural machine translation by jointly learning to align and
  translate.
\newblock \emph{arXiv preprint arXiv:1409.0473}, 2014.

\bibitem[Barone et~al.(2017)Barone, Helcl, Sennrich, Haddow, and
  Birch]{barone2017deep}
Antonio Valerio~Miceli Barone, Jind{\v{r}}ich Helcl, Rico Sennrich, Barry
  Haddow, and Alexandra Birch.
\newblock Deep architectures for neural machine translation.
\newblock In \emph{Proceedings of the Second Conference on Machine
  Translation}, pp.\  99--107, 2017.

\bibitem[Bastings et~al.(2019)Bastings, Aziz, and Titov]{bastings2019}
Joost Bastings, Wilker Aziz, and Ivan Titov.
\newblock Interpretable neural predictions with differentiable binary
  variables.
\newblock \emph{Proceedings of the 57th Annual Meeting of the Association for
  Computational Linguistics (Volume 1: Long Papers)}, pp.\  2963--2973, 2019.

\bibitem[Bengio et~al.(2013)Bengio, L{\'e}onard, and
  Courville]{bengio2013estimating}
Yoshua Bengio, Nicholas L{\'e}onard, and Aaron Courville.
\newblock Estimating or propagating gradients through stochastic neurons for
  conditional computation.
\newblock \emph{arXiv preprint arXiv:1308.3432}, 2013.

\bibitem[Bottou \& Cun(2004)Bottou and Cun]{BottouEtAl2004}
L\'{e}on Bottou and Yann~L. Cun.
\newblock Large scale online learning.
\newblock In S.~Thrun, L.~K. Saul, and B.~Sch\"{o}lkopf (eds.), \emph{Advances
  in Neural Information Processing Systems 16}, pp.\  217--224. MIT Press,
  2004.

\bibitem[Boudchiche et~al.(2017)Boudchiche, Mazroui, Bebah, Lakhouaja, and
  Boudlal]{boudchiche2017alkhalil}
Mohamed Boudchiche, Azzeddine Mazroui, Mohamed Ould Abdallahi~Ould Bebah,
  Abdelhak Lakhouaja, and Abderrahim Boudlal.
\newblock Alkhalil morpho sys 2: A robust arabic morpho-syntactic analyzer.
\newblock \emph{Journal of King Saud University-Computer and Information
  Sciences}, 29\penalty0 (2):\penalty0 141--146, 2017.

\bibitem[Cettolo(2012)]{mauro2012wit3}
Mauro Cettolo.
\newblock Wit3: Web inventory of transcribed and translated talks.
\newblock In \emph{Conference of European Association for Machine Translation},
  pp.\  261--268, 2012.

\bibitem[Cherry et~al.(2018)Cherry, Foster, Bapna, Firat, and
  Macherey]{cherry2018revisiting}
Colin Cherry, George Foster, Ankur Bapna, Orhan Firat, and Wolfgang Macherey.
\newblock Revisiting character-based neural machine translation with capacity
  and compression.
\newblock In \emph{Proceedings of the 2018 Conference on Empirical Methods in
  Natural Language Processing}, pp.\  4295--4305, 2018.

\bibitem[Cho et~al.(2014)Cho, van Merrienboer, Bahdanau, and
  Bengio]{cho2014properties}
Kyunghyun Cho, Bart van Merrienboer, Dzmitry Bahdanau, and Yoshua Bengio.
\newblock On the properties of neural machine translation: Encoder--decoder
  approaches.
\newblock In \emph{Proceedings of 8th Workshop on Syntax, Semantics and
  Structure in Statistical Translation (SSST)}, pp.\  103--111, 2014.

\bibitem[Clark et~al.(2011)Clark, Dyer, Lavie, and Smith]{clark2011better}
Jonathan~H Clark, Chris Dyer, Alon Lavie, and Noah~A Smith.
\newblock Better hypothesis testing for statistical machine translation:
  Controlling for optimizer instability.
\newblock In \emph{Proceedings of the 49th Annual Meeting of the Association
  for Computational Linguistics (Volume 2: Short Papers)}, pp.\  176--181.
  Association for Computational Linguistics, 2011.

\bibitem[Cotterell et~al.(2018)Cotterell, Mielke, Eisner, and
  Roark]{cotterell-etal-2018-languages}
Ryan Cotterell, Sebastian~J. Mielke, Jason Eisner, and Brian Roark.
\newblock Are all languages equally hard to language-model?
\newblock In \emph{Proceedings of the 2018 Conference of the North {A}merican
  Chapter of the Association for Computational Linguistics: Human Language
  Technologies (Volume 2: Short Papers)}, pp.\  536--541, 2018.

\bibitem[Cuong \& Simaan(2014)Cuong and Simaan]{cuong2014latent}
Hoang Cuong and Khalil Simaan.
\newblock Latent domain translation models in mix-of-domains haystack.
\newblock In \emph{Proceedings of COLING 2014, the 25th International
  Conference on Computational Linguistics}, pp.\  1928--1939, 2014.

\bibitem[Goyal et~al.(2017)Goyal, Sordoni, C{\^o}t{\'e}, Ke, and
  Bengio]{goyal2017z}
Anirudh Goyal Alias~Parth Goyal, Alessandro Sordoni, Marc-Alexandre
  C{\^o}t{\'e}, Nan~Rosemary Ke, and Yoshua Bengio.
\newblock Z-forcing: Training stochastic recurrent networks.
\newblock In \emph{Advances in neural information processing systems}, pp.\
  6713--6723, 2017.

\bibitem[Jang et~al.(2017)Jang, Gu, and Poole]{JangEtAl2017:GumbelSoftmax}
Eric Jang, Shixiang Gu, and Ben Poole.
\newblock Categorical reparameterization with {G}umbel-{S}oftmax.
\newblock \emph{International Conference on Learning Representations}, 2017.

\bibitem[Jordan et~al.(1999)Jordan, Ghahramani, Jaakkola, and
  Saul]{Jordan+1999:VI}
MichaelI. Jordan, Zoubin Ghahramani, TommiS. Jaakkola, and Lawrence~K. Saul.
\newblock An introduction to variational methods for graphical models.
\newblock \emph{Machine Learning}, 37\penalty0 (2):\penalty0 183--233, 1999.

\bibitem[Kinga \& Ba(2014)Kinga and Ba]{kinga2015method}
D~Kinga and J~Ba.
\newblock Adam: A method for stochastic optimization.
\newblock \emph{arXiv preprint arXiv:1412.6980}, 2014.

\bibitem[Kingma \& Welling(2013)Kingma and Welling]{Kingma+2014:VAE}
Diederik~P. Kingma and Max Welling.
\newblock Auto-encoding variational {B}ayes.
\newblock \emph{arXiv preprint arXiv:1312.6114}, 2013.

\bibitem[Klein et~al.(2017)Klein, Kim, Deng, Senellart, and
  Rush]{klein2017opennmt}
Guillaume Klein, Yoon Kim, Yuntian Deng, Jean Senellart, and Alexander Rush.
\newblock Open{N}{M}{T}: Open-source toolkit for neural machine translation.
\newblock \emph{Proceedings of the 55th Annual Meeting of the Association for
  Computational Linguistics, System Demonstrations}, pp.\  67--72, 2017.

\bibitem[Kreutzer \& Sokolov(2018)Kreutzer and Sokolov]{kreutzer2018learning}
Julia Kreutzer and Artem Sokolov.
\newblock Learning to segment inputs for nmt favors character-level processing.
\newblock In \emph{Proceedings of the 15th International Workshop on Spoken
  Language Translation}, pp.\  166--172, 2018.

\bibitem[Kumaraswamy(1980)]{kumaraswamy1980generalized}
Ponnambalam Kumaraswamy.
\newblock A generalized probability density function for double-bounded random
  processes.
\newblock \emph{Journal of Hydrology}, 46\penalty0 (1-2):\penalty0 79--88,
  1980.

\bibitem[Ling et~al.(2015)Ling, Trancoso, Dyer, and
  Black]{DBLP:journals/corr/LingTDB15}
Wang Ling, Isabel Trancoso, Chris Dyer, and Alan~W. Black.
\newblock Character-based neural machine translation.
\newblock \emph{arXiv preprint arXiv:1511.04586}, 2015.

\bibitem[Lison \& Tiedemann(2016)Lison and
  Tiedemann]{lison2016opensubtitles2016}
Pierre Lison and J{\"o}rg Tiedemann.
\newblock Opensubtitles2016: Extracting large parallel corpora from movie and
  {T}{V} subtitles.
\newblock 2016.

\bibitem[Louizos et~al.(2018)Louizos, Welling, and Kingma]{louizos2017learning}
Christos Louizos, Max Welling, and Diederik~P Kingma.
\newblock Learning sparse neural networks through {L}$_0$ regularization.
\newblock \emph{arXiv preprint arXiv:1712.01312}, 2018.

\bibitem[Luong \& Manning(2016)Luong and Manning]{luong-hybrid}
Minh-Thang Luong and Christopher~D. Manning.
\newblock Achieving open vocabulary neural machine translation with hybrid
  word-character models.
\newblock In \emph{Proceedings of the 54th Annual Meeting of the Association
  for Computational Linguistics (Volume 1: Long Papers)}, pp.\  1054--1063,
  2016.

\bibitem[Luong et~al.(2015)Luong, Pham, and Manning]{luong2015effective}
Minh-Thang Luong, Hieu Pham, and Christopher~D Manning.
\newblock Effective approaches to attention-based neural machine translation.
\newblock In \emph{Proceedings of the 2015 Conference on Empirical Methods in
  Natural Language Processing}, pp.\  1412--1421, 2015.

\bibitem[Maddison et~al.(2017)Maddison, Mnih, and
  Teh]{MaddisonEtAl2017:Concrete}
Chris~J. Maddison, Andriy Mnih, and Yee~Whye Teh.
\newblock The concrete distribution: A continous relaxation of discrete random
  variables.
\newblock \emph{International Conference on Learning Representations}, 2017.

\bibitem[Nalisnick \& Smyth(2016)Nalisnick and Smyth]{nalisnick2017stick}
Eric Nalisnick and Padhraic Smyth.
\newblock Stick-breaking variational autoencoders.
\newblock \emph{arXiv preprint arXiv:1605.06197}, 2016.

\bibitem[Oflazer \& Kuru{\"o}z(1994)Oflazer and Kuru{\"o}z]{oflazer1994tagging}
Kemal Oflazer and Ilker Kuru{\"o}z.
\newblock Tagging and morphological disambiguation of turkish text.
\newblock In \emph{Proceedings of the fourth conference on Applied natural
  language processing}, pp.\  144--149. Association for Computational
  Linguistics, 1994.

\bibitem[Papineni et~al.(2002)Papineni, Roukos, Ward, and Zhu]{bleu}
Kishore Papineni, Salim Roukos, Todd Ward, and Wei-Jing Zhu.
\newblock {BLEU}: a {M}ethod for {A}utomatic {E}valuation of {M}achine
  {T}ranslation.
\newblock In \emph{Proceedings of the 40th Annual Meeting of the Association
  for Computational Linguistics}, pp.\  311--318, 2002.

\bibitem[Paszke et~al.(2017)Paszke, Gross, Chintala, Chanan, Yang, DeVito, Lin,
  Desmaison, Antiga, and Lerer]{paszke2017automatic}
Adam Paszke, Sam Gross, Soumith Chintala, Gregory Chanan, Edward Yang, Zachary
  DeVito, Zeming Lin, Alban Desmaison, Luca Antiga, and Adam Lerer.
\newblock Automatic differentiation in {P}ytorch.
\newblock \emph{NeurIPS Autodiff Workshop}, 2017.

\bibitem[Popovi{\'c}(2015)]{popovic2015chrf}
Maja Popovi{\'c}.
\newblock chrf: character n-gram f-score for automatic mt evaluation.
\newblock In \emph{Proceedings of the Tenth Workshop on Statistical Machine
  Translation}, pp.\  392--395, 2015.

\bibitem[Rezende et~al.(2014)Rezende, Mohamed, and Wierstra]{RezendeEtAl14VAE}
Danilo~Jimenez Rezende, Shakir Mohamed, and Daan Wierstra.
\newblock Stochastic backpropagation and approximate inference in deep
  generative models.
\newblock In \emph{Proceedings of the 31st International Conference on Machine
  Learning}, volume~32 of \emph{Proceedings of Machine Learning Research}, pp.\
   1278--1286, 2014.

\bibitem[Robbins \& Monro(1951)Robbins and Monro]{robbinsmonro:1951}
Herbert Robbins and Sutton Monro.
\newblock A stochastic approximation method.
\newblock \emph{The Annals of Mathematical Statistics}, 22\penalty0
  (3):\penalty0 400--407, 1951.

\bibitem[Sahin \& Steedman(2018)Sahin and Steedman]{sahin2018character}
Gozde~Gul Sahin and Mark Steedman.
\newblock Character-level models versus morphology in semantic role labeling.
\newblock In \emph{Proceedings of the 56th Annual Meeting of the Association
  for Computational Linguistics (Volume 1: Long Papers)}, pp.\  386--396, 2018.

\bibitem[Sak et~al.(2007)Sak, G{\"u}ng{\"o}r, and
  Sara{\c{c}}lar]{sak2007morphological}
Ha{\c{s}}im Sak, Tunga G{\"u}ng{\"o}r, and Murat Sara{\c{c}}lar.
\newblock Morphological disambiguation of turkish text with perceptron
  algorithm.
\newblock In \emph{International Conference on Intelligent Text Processing and
  Computational Linguistics}, pp.\  107--118. Springer, 2007.

\bibitem[Schulz et~al.(2018)Schulz, Aziz, and
  Cohn]{schulz-etal-2018-stochastic}
Philip Schulz, Wilker Aziz, and Trevor Cohn.
\newblock A stochastic decoder for neural machine translation.
\newblock In \emph{Proceedings of the 56th Annual Meeting of the Association
  for Computational Linguistics (Volume 1: Long Papers)}, pp.\  1243--1252,
  2018.

\bibitem[Sennrich(2017)]{sennrich2017grammatical}
Rico Sennrich.
\newblock How grammatical is character-level neural machine translation?
  {A}ssessing {M}{T} quality with contrastive translation pairs.
\newblock In \emph{Proceedings of the 15th Conference of the European Chapter
  of the Association for Computational Linguistics (Volume 2, Short Papers)},
  pp.\  376--382, 2017.

\bibitem[Sennrich et~al.(2016)Sennrich, Haddow, and Birch]{sennrich2016neural}
Rico Sennrich, Barry Haddow, and Alexandra Birch.
\newblock Neural machine translation of rare words with subword units.
\newblock In \emph{Proceedings of the 54th Annual Meeting of the Association
  for Computational Linguistics (Volume 1: Long Papers)}, pp.\  1715--1725,
  2016.

\bibitem[Skadi{\c{n}}{\v{s}} et~al.(2014)Skadi{\c{n}}{\v{s}}, Tiedemann, Rozis,
  and Deksne]{skadicnvs2014billions}
Raivis Skadi{\c{n}}{\v{s}}, J{\"o}rg Tiedemann, Roberts Rozis, and Daiga
  Deksne.
\newblock Billions of parallel words for free: Building and using the eu
  bookshop corpus.
\newblock In \emph{Proceedings of the Ninth International Conference on
  Language Resources and Evaluation (LREC)}, pp.\  1850--1855, 2014.

\bibitem[Smith(2011)]{smith2011linguistic}
Noah~A Smith.
\newblock Linguistic structure prediction.
\newblock \emph{Synthesis lectures on human language technologies}, 4\penalty0
  (2):\penalty0 1--274, 2011.

\bibitem[Strakov\'{a} et~al.(2014)Strakov\'{a}, Straka, and
  Haji\v{c}]{strakova14}
Jana Strakov\'{a}, Milan Straka, and Jan Haji\v{c}.
\newblock Open-{S}ource {T}ools for {M}orphology, {L}emmatization, {POS}
  {T}agging and {N}amed {E}ntity {R}ecognition.
\newblock In \emph{Proceedings of 52nd Annual Meeting of the Association for
  Computational Linguistics: System Demonstrations}, pp.\  13--18. Association
  for Computational Linguistics, 2014.

\bibitem[Tiedemann(2009)]{tiedemann2009news}
J{\"o}rg Tiedemann.
\newblock News from opus-a collection of multilingual parallel corpora with
  tools and interfaces.
\newblock In \emph{Recent advances in natural language processing}, volume~5,
  pp.\  237--248, 2009.

\bibitem[Tiedemann(2012)]{tiedemann2012parallel}
J{\"o}rg Tiedemann.
\newblock Parallel data, tools and interfaces in opus.
\newblock In \emph{Proceedings of the Seventh International Conference on
  Language Resources and Evaluation (LREC)}, pp.\  2214--2218, 2012.

\bibitem[Tyers \& Alperen(2010)Tyers and Alperen]{tyers2010south}
Francis~M Tyers and Murat~Serdar Alperen.
\newblock South-east european times: A parallel corpus of balkan languages.
\newblock In \emph{Proceedings of the LREC Workshop on Exploitation of
  Multilingual Resources and Tools for Central and (South-) Eastern European
  Languages}, pp.\  49--53, 2010.

\bibitem[Vania \& Lopez(2017)Vania and Lopez]{vania2017characters}
Clara Vania and Adam Lopez.
\newblock From characters to words to in between: Do we capture morphology?
\newblock In \emph{Proceedings of the 55th Annual Meeting of the Association
  for Computational Linguistics (Volume 1: Long Papers)}, pp.\  2016--2027,
  2017.

\bibitem[Williams(1992)]{Williams1992}
Ronald~J Williams.
\newblock Simple statistical gradient-following algorithms for connectionist
  reinforcement learning.
\newblock \emph{Machine learning}, 8\penalty0 (3-4):\penalty0 229--256, 1992.

\bibitem[Wu et~al.(2016)Wu, Schuster, Chen, Le, Norouzi, Macherey, Krikun, Cao,
  Gao, Macherey, et~al.]{wu2016google}
Yonghui Wu, Mike Schuster, Zhifeng Chen, Quoc~V Le, Mohammad Norouzi, Wolfgang
  Macherey, Maxim Krikun, Yuan Cao, Qin Gao, Klaus Macherey, et~al.
\newblock Google’s neural machine translation system: Bridging the gap
  between human and machine translation.
\newblock 2016.

\bibitem[Zhang et~al.(2016)Zhang, Xiong, Su, Duan, and
  Zhang]{zhang-etal-2016-variational-neural}
Biao Zhang, Deyi Xiong, Jinsong Su, Hong Duan, and Min Zhang.
\newblock Variational neural machine translation.
\newblock In \emph{Proceedings of the 2016 Conference on Empirical Methods in
  Natural Language Processing}, pp.\  521--530, 2016.

\bibitem[Zhou \& Neubig(2017)Zhou and Neubig]{zhou2017multi}
Chunting Zhou and Graham Neubig.
\newblock Multi-space variational encoder-decoders for semi-supervised labeled
  sequence transduction.
\newblock In \emph{Proceedings of the 55th Annual Meeting of the Association
  for Computational Linguistics (Volume 1: Long Papers)}, pp.\  310--320, 2017.

\end{thebibliography}
\bibliographystyle{iclr2020_conference}

\clearpage
\appendix
\section{Appendix}

\subsection{The statistical characteristics of experimental data}

\begin{table}[h!]
  \begin{centering}
    \begin{tabular}{c|c|cc|cc}
    \textbf{Language Pair} & \textbf{\# sentences} & \multicolumn{2}{c|}{\textbf{\# tokens}} & \multicolumn{2}{c}{\textbf{\# types}}\\
     & & {\bf Source} & {\bf Target} & {\bf Source} & {\bf Target} \\
    \hline
     English-Arabic & 238K & 5M & 4M & 120K & 220K \\
    \hline
     English-Czech & 118K & 2M & 2M & 50K & 118K \\
    \hline
    English-Turkish & 136K & 2M & 3M & 53K & 171K \\
    \end{tabular}
    \caption{Training sets based on the TED Talks corpora (\textit{M}: Million, \textit{K}: Thousand).}
    \label{tab:data1}
  \end{centering}
\end{table}

\begin{table}[h!]
  \begin{centering}
    \begin{tabular}{c|c|cc|cc}
    \textbf{Language Pair} & \textbf{\# sentences} & \multicolumn{2}{c|}{\textbf{\# tokens}} & \multicolumn{2}{c}{\textbf{\# types}}\\
     & & {\bf Source} & {\bf Target} & {\bf Source} & {\bf Target} \\
    \hline
    English-Turkish & 434K & 8M & 6M & 135K & 373K\\
    \end{tabular}
	\caption{The multi-domain training set (\textit{M}: Million, \textit{K}: Thousand).}
  \label{tab:data2}
  \end{centering}
\end{table}

 \begin{table}[h!]
    \begin{centering}
    \begin{tabular}{c|c|c|c}
    \textbf{Language} &\multicolumn{2}{c|}{\textbf{Data sets }} &\textbf{\# sentences}  \\
    \hline
    English-Arabic & Development & dev2010, test2010 & 6K \\
    & &test2011, test2012 & \\
    &  Testing & test2013, test2014 & 4K\\
    \hline
    English-Czech & Development & dev2010, test2010, & 3K \\
    & &test2011 &  \\
    &  Testing & test2012, test2013& 3K \\
    \hline
    English-Turkish & Development & dev2010, test2010& 3K  \\
    &  Testing & test2011, test2012& 3K\\
    \end{tabular}
    \caption{Development and testing sets (\textit{K}: Thousand). }
    \label{tab:devtest}
  \end{centering}
\end{table}

\subsection{The Kumaraswamy distribution}

\begin{figure}[!h]
\centering
    \includegraphics[scale=0.5]{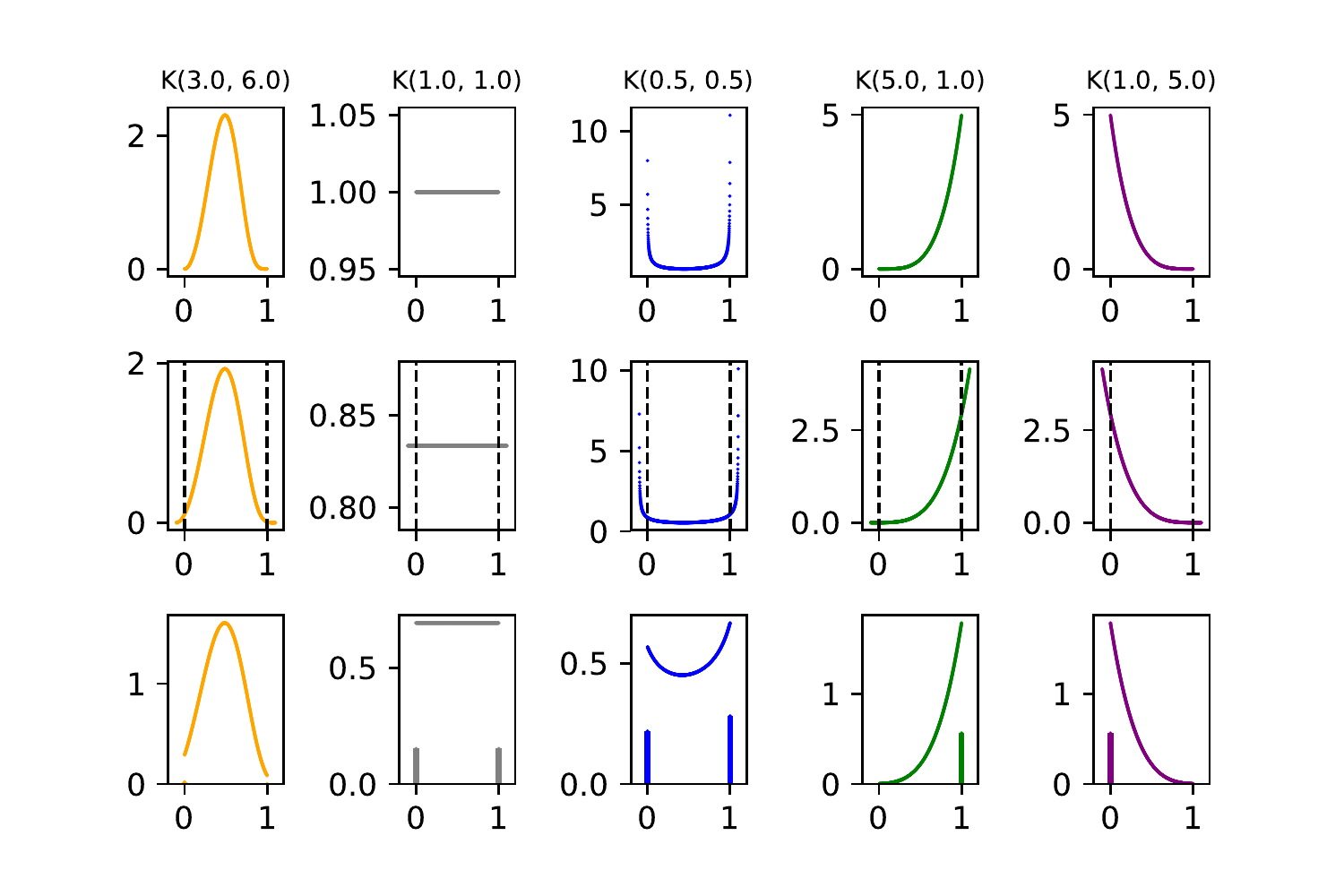}
    \caption{\label{fig:kuma}The top row shows the density function of the continuous base distribution over $(0,1)$. The middle row shows the result of stretching it to include $0$ and $1$ in its support. The bottom row shows the result of rectification: probability mass under $(l,0)$ collapses to $0$ and probability mass under $(1,r)$ collapses to $1$, which cause sparse outcomes to have non-zero mass. Varying the shape parameters $(a,b)$ of the underlying continuous distribution changes how much mass concentrates outside the support $(0, 1)$ in the stretched density, and hence the probability of sampling sparse outcomes.}
\end{figure}

\subsection{The Effect of Feature Dimensions}
\label{subsec:featuredim}

\begin{figure}[h]
    \centering
    \includegraphics[scale=0.4]{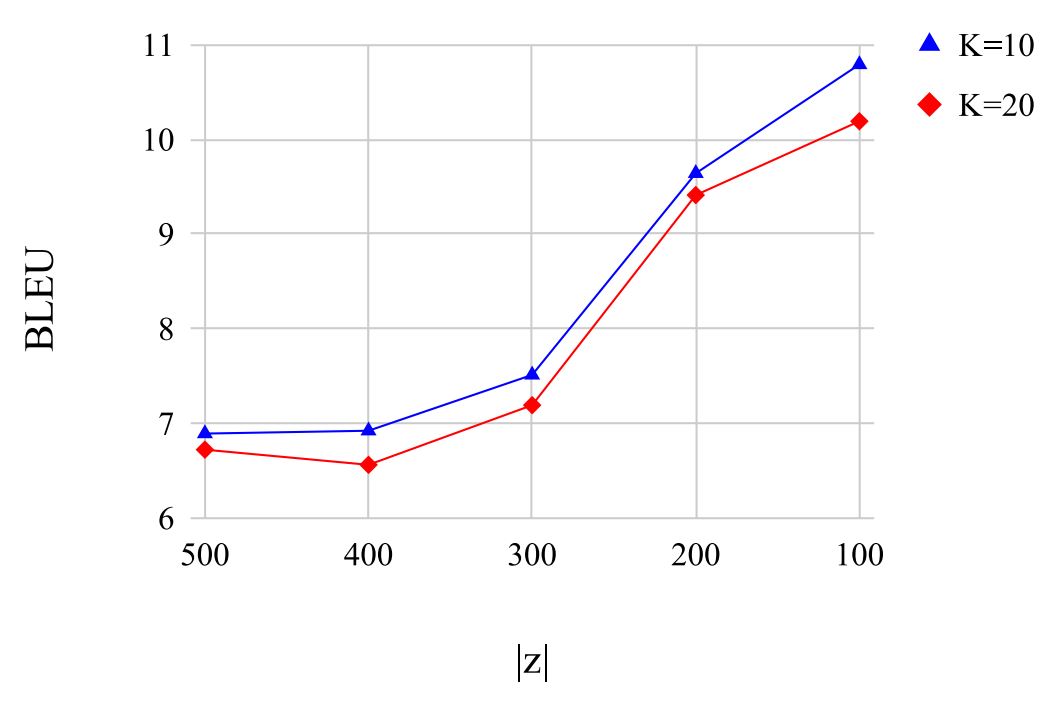}
    \caption{The effect of feature dimensions on translation accuracy in Turkish.}
    \label{fig:featuredim}
\end{figure}

We investigate the optimal lemma and inflectional feature sizes by measuring the accuracy in English-to-Turkish translation using different feature vector dimensions. The results given in Figure \ref{fig:featuredim} show that gradually compressing the word representations computed by recurrent hidden states, with an original dimension of 512, from 500 to 100, leads to increased output accuracy, suggesting that encoding more compact representations might provide the model with a better generalization capability. Our results also show that using a feature dimension of 10 is sufficient in reaching the best accuracy.

\end{document}